\documentclass{article}
\usepackage{spconf,amsmath,graphicx,hyperref}

\usepackage{array}

\usepackage{cuted} 
\usepackage{caption} 
\usepackage{lipsum}  
\usepackage{amsfonts} 
\usepackage{booktabs} 
\usepackage{multirow}
\usepackage{subcaption}
\usepackage{graphicx}     
\newlength\savewidth
\newcommand\shline{\noalign{\global\savewidth\arrayrulewidth
                           \global\arrayrulewidth 1pt}%
                           \hline \noalign{\global\arrayrulewidth\savewidth}}

\title{RANKING-AWARE REINFORCEMENT LEARNING FOR ORDINAL RANKING}
\name{Aiming Hao, Chen Zhu, Jiashu Zhu, Jiahong Wu\sthanks{Corresponding Author}, Xiangxiang Chu}
\address{AMAP, Alibaba Group}
\begin{document}
%
\maketitle
\begin{abstract}
Ordinal regression and ranking are challenging due to inherent ordinal dependencies that conventional methods struggle to model. We propose Ranking-Aware Reinforcement Learning (RARL), a novel RL framework that explicitly learns these relationships. At its core, RARL features a unified objective that synergistically integrates regression and Learning-to-Rank (L2R), enabling mutual improvement between the two tasks. This is driven by a ranking-aware verifiable reward that jointly assesses regression precision and ranking accuracy, facilitating direct model updates via policy optimization. To further enhance training, we introduce Response Mutation Operations (RMO), which inject controlled noise to improve exploration and prevent stagnation at saddle points. The effectiveness of RARL is validated through extensive experiments on three distinct benchmarks.
\end{abstract}
\section{INTRODUCTION}
Ordinal regression and ranking are fundamental tasks in signal processing and machine learning, where the objective is to predict and sort data based on labels with inherent order. These tasks appear in diverse applications, from medical signal analysis (e.g., severity staging)~\cite{cheng2023robust, yu2024clip} to content quality assessment (e.g., audio/image aesthetics)~\cite{wang2023learning, kong2016photo}. Unlike standard regression or classification, ordinal tasks demand that a model captures both the continuous nature of values and the hierarchical dependencies among categories. This dual requirement poses a significant challenge: models must balance regression accuracy (predicting exact values) and ranking consistency (preserving relative order).

Conventional methods often fail to address this trade-off effectively. Regression models ~\cite{shen2017label, rothe2018deep, dai2023semi} minimize point-wise errors but neglect ordinal dependencies, while classification-based approaches~\cite{geng2013facial, diaz2019soft} lose granularity by discretizing labels. Even Learning-to-Rank (L2R) methods~\cite{niu2016ordinal, chen2017using}, which focus on relative order, struggle to jointly optimize both objectives within a single framework. Crucially, these paradigms lack an explicit mechanism to model the causal relationships between input signals and their ordinal structure.

Recent advances have shown Reinforcement Learning (RL)~\cite{liu2025understanding, chu2025gpg} to be a powerful tool for optimizing complex, multi-objective rewards, particularly in the context of Large Language Models (LLMs)~\cite{guo2025deepseek}. Inspired by this, we propose Ranking-Aware Reinforcement Learning (RARL), a novel RL framework that directly models ordinal dependencies. RARL unifies regression and L2R through a ranking-aware verifiable reward, which allows a policy to be optimized for both regression accuracy and ranking consistency simultaneously. Furthermore, to combat the entropy collapse phenomenon~\cite{yu2025dapo} common in RL, we introduce the Response Mutation Operation (RMO), a mechanism to enhance exploration and escape saddle points. Our contributions are: 

• A unified RL framework RARL is introduced to jointly optimize regression and ranking via ranking-aware verifiable rewards, ensuring bidirectional alignment between objectives.  

• RMO is developed to alleviate the entropy collapse phenomenon during GRPO training and reactivate gradient signals to escape the saddle points, thereby motivating further exploration of evolutionary strategies in RL frameworks.

Finally, we demonstrate through extensive experiments that RARL achieves significantly better performance on three distinct benchmarks: facial age estimation~\cite{zhang2017age}, object count ranking~\cite{singh2024benchmarking}, and aesthetic assessment~\cite{ghosal2019aesthetic}, proving its effectiveness and generalizability.

\section{BACKGROUND}
\noindent \textbf{Reinforcement Learning Fine-tuning}. Our work is grounded in the Reinforcement Learning with Verifiable Rewards (RLVR) paradigm~\cite{guo2025deepseek, team2025kimi}, which refines a policy model $\pi_{\theta}$ by optimizing for a task-specific, verifiable reward function $R:\left(\boldsymbol{q},\boldsymbol{o}\right)\mapsto \mathbb{R}$ instead of a learned reward model. This approach eliminates reward model bias and is ideal for tasks with clear correctness criteria. The objective maximizes the expected reward, regularized by a KL-divergence term to prevent large deviations from a reference policy $\pi_{\text{ref}}$:
\begin{equation}
\begin{aligned}
\max_{\theta}\ \mathbb{E}_{\boldsymbol{q},\boldsymbol{o}\sim \pi_{\theta}}\left[R\left(\boldsymbol{q},\boldsymbol{o}\right)\right]-\beta \mathbb{E}_{\boldsymbol{q}}\left[\text{KL}\left(\pi_{\theta}\left(\boldsymbol{o}\mid \boldsymbol{q}\right)\parallel \pi_{\text{ref}}\left(\boldsymbol{o}\mid \boldsymbol{q}\right)\right)\right],
\end{aligned}
\end{equation}
We specifically employ Group Relative Policy Optimization (GRPO), an advanced RLVR algorithm. For each input $\boldsymbol{q}$, GRPO samples a group of responses $\left\{\boldsymbol{o}_i\right\}_{i=1}^G$ and calculates a normalized, intra-group advantage $\hat{A}_{i}$ for each response:
\begin{equation}
\begin{aligned}
\hat{A}_{i}=\tilde{r}_i=\frac{r_i-\text{mean}\left(\left\{r_i\right\}_{i=1}^G\right)}{\text{std}\left(\left\{r_i\right\}_{i=1}^G\right)}.
\end{aligned}
\end{equation}
This relative advantage encourages the model to prioritize responses that outperform others within the same generation group. Our proposed framework leverages this mechanism with custom-designed rewards for ordinal ranking.

\begin{table}[t]
\caption{
\small
\textbf{Prompts used in RARL.} We have listed the system and instruct prompt separately.}
\vspace{-4mm}
\label{tab:prompt}
\begin{center}
\setlength{\tabcolsep}{4pt}
\scalebox{0.9}
{
\begin{tabular}{p{9cm}}  
\toprule
\textbf{System Prompt:} The user asks a question, and the Assistant solves it. The assistant first thinks about the reasoning process in the mind and then provides the user with the answer. The reasoning process and answer are enclosed within $<$think$>$ $<$/think$>$ and $<$answer$>$ $<$/answer$>$ tags, respectively, i.e., $<$think$>$ reasoning process here $<$/think$>$ $<$answer$>$ answer here $<$/answer$>$.
\\
\midrule
\textbf{Instruct Prompt:} Analyze each image to determine the [Rule], and if there are multiple images, sort them in ascending order of [Rule]. Output the results as a JSON list of dictionaries in this format: [\{'image\_id': number, 'value': value\}, ...], where value is the [Rule] ([Value\_Range]). Ensure the list is sorted correctly with strictly followed syntax.
\\
\bottomrule
\end{tabular}
}
\vspace{-20pt}
\end{center}
\end{table}

\noindent \textbf{Problem Formulation}. Ordinal regression is a structured prediction task where labels possess inherent ordinal relationships, such as facial age levels or aesthetic scores. Extending this concept, the ordinal ranking task dynamically sequences images by fusing label hierarchies (e.g., age progression) with textual semantics (e.g., rank from youngest to oldest). These two tasks can be unified under an input-dependent framework: single-image inputs trigger pure ordinal regression, while multi-image batches require simultaneous optimization of regression accuracy and ranking consistency through multi-task learning. Mathematically, given an input text prompt $\boldsymbol{q}$ and an unordered list of $N$ images $\boldsymbol{X}=\left[\boldsymbol{x}_1,\cdots \boldsymbol{x}_{N}\right]$, each image $\boldsymbol{x}_{i}$ is assigned an ordinal rank $y_{i} \in \left\{r_1, ..., r_M\right\}$ under $\boldsymbol{q}$, where $r_M \succ r_{M-1} \succ \cdots \succ r_1$ defines a total order over $M$ distinct ranks. Here, $\succ$ denotes strict precedence (e.g., higher age level or quality). The model $\mathcal{M}$ predicts the ordinal ranks $\tilde{\boldsymbol{Y}}=\left\{\tilde{\boldsymbol{y}}_i\right\}_{i=1}^N$ and a permutation $\pi$ that sorts $\boldsymbol{X}$ into an ordered sequence $\tilde{\boldsymbol{X}}=\left[\boldsymbol{x}_{\pi\left(1\right)},\cdots \boldsymbol{x}_{\pi\left(N\right)}\right]$ satisfying $\boldsymbol{y}_{\pi\left(1\right)}\succ \cdots \succ \boldsymbol{y}_{\pi\left(N\right)}$. To model the above tasks, we implement $\mathcal{M}$ as a VLM that generates JSON-formatted outputs (prompt templates in Table~\ref{tab:prompt}) with two structured fields: 

• \textit{value}: Predicted ordinal label; 

• \textit{image\_id}: Image IDs in ranked order.

\noindent Then, the RARL framework is proposed to unify ordinal image regression and ranking under a reinforcement learning paradigm, where verifiable reward functions assess both the accuracy of the regression values and the correctness of the ranking permutation, as detailed in the next section.
\section{METHODOLOGY}
\subsection{Ranking-Aware Verifiable Rewards}
\label{sec:rewards}

\begin{figure*}[t]
\begin{center}
\includegraphics[width=0.9\linewidth]{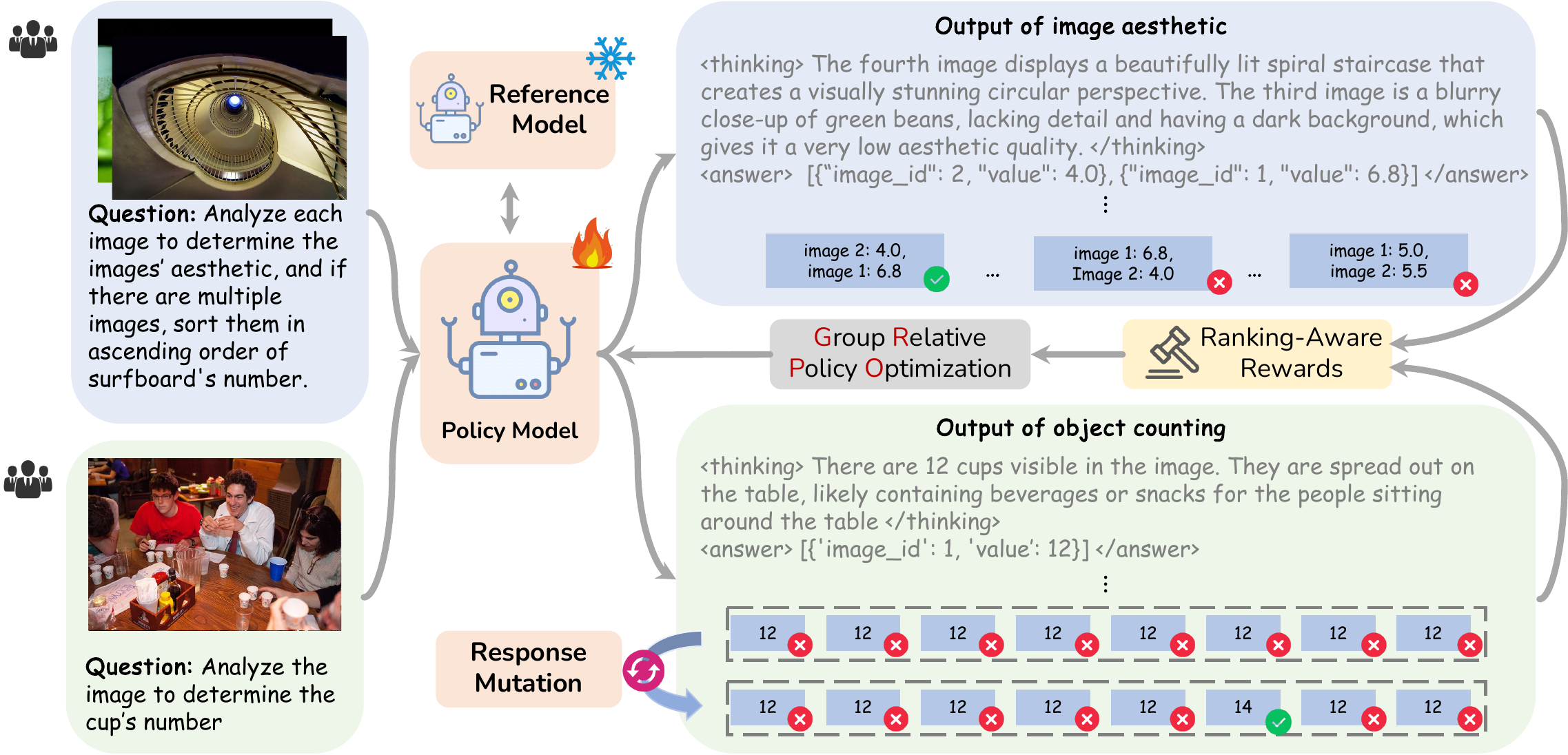}
\end{center}
\vspace{-10pt}
\caption{Flowchart of proposed RARL. For a given image-question pair, our RARL model generates multiple reasoning-based responses. It is then optimized via policy gradient using ranking-aware verifiable rewards. Then, the response mutation operation is employed to reactivate gradient signals and escape saddle points.}
\label{fig:pipeline}
\vspace{-10pt}	
\end{figure*}

\noindent The structure of the proposed RARL is presented in Fig. ~\ref{fig:pipeline}. In RARL, instead of a learned reward model, we employ a verifiable reward function to quantify prediction correctness. We design this function with two primary components: a Regression Reward and a Ranking Reward.

\noindent \textbf{Regression Reward in Ordinal Regression Tasks}. 
To measure the accuracy of the predicted continuous value $\tilde{\boldsymbol{y}}_i$, we use an L2-inspired reward that is non-zero only if the prediction is within a tolerance $\delta$ of the ground truth $\boldsymbol{y}_i$. It is defined as:
\begin{equation}
\begin{aligned}
r^{\text{reg}}_i=\begin{cases}
\left(1 - \frac{\left|\tilde{\boldsymbol{y}}_i-\boldsymbol{y}_i\right|}{2\delta}\right)^2 \text{, if}\ \ \boldsymbol{y}_i-\delta\leq \tilde{\boldsymbol{y}}_i\leq \boldsymbol{y}_i+\delta
\\
0 \ \ \ \ \ \ \ \ \ \  \ \ \ \  \ \ \ \  \ \ \ \ \ \text{, otherwise}
\end{cases}
\end{aligned}
\end{equation}
The regression reward $R_{\text{reg}}$ is the mean over all instances.

\noindent \textbf{Ranking Reward for Ordinal Ranking Tasks}. This composite reward assesses the quality of the predicted permutation $\tilde{\pi}$ and is a sum of three terms: $R_{\text{rank}}=r^{\text{len}}+r^{\text{consis}}+r^{\text{acc}}$.


1) Length Compliance Reward---This reward ensures the predicted sequence length aligns with the ground truth, thereby satisfying application constraints:
\begin{equation}
\begin{aligned}
r^{\text{len}}=\begin{cases}
1 - \frac{\#\pi-\#\tilde{\pi}}{\#\pi} \text{, if}\ \ \#\tilde{\pi} \leq \#\pi
\\
0 \ \ \ \ \ \ \ \ \ \  \ \ \ \ \ \ \text{, otherwise}
\end{cases},
\end{aligned}
\end{equation}
where $\#{\pi}$ denotes the length of $\pi$.

2) Ranking Consistency Reward---To enforce coherence between explicit ranking predictions $\tilde{\pi}$ and implicit order $\tilde{\pi}_{reg}$ derived from regression outputs, this reward is defined as:
\begin{equation}
\begin{aligned}
r^{\text{consis}}=\left(\tau\left(\tilde{\pi},\tilde{\pi}_{reg}\right)+1\right)/2,
\end{aligned}
\end{equation}
where $\tau$ represents Kendall’s Tau coefficient.

3) Ranking Accuracy Reward---To optimize ranking agreement with human annotations or reference standards directly, this reward is computed as:
\begin{equation}
\begin{aligned}
r^{\text{acc}}=\left(\tau\left(\tilde{\pi},\pi\right)+1\right)/2.
\end{aligned}
\end{equation}

The composite ranking reward $R_{\text{rank}}$ combines these objectives via linear summation:
\begin{equation}
\begin{aligned}
R_{\text{rank}}=r^{\text{len}}+r^{\text{consis}}+r^{\text{acc}}.
\end{aligned}
\end{equation}
This unified formulation ensures the model balances length control, internal consistency, and human alignment.

\noindent \textbf{Ranking-Aware Reward}. The final Ranking-Aware Reward $R_{\text{final}}$ linearly combines three rewards:
\begin{equation}
\begin{aligned}
R_{\text{final}}=\lambda_1 R_{\text{reg}}+\lambda_2 R_{\text{rank}}+\lambda_3 R_{\text{format}}
\end{aligned}
\end{equation}
where $R_{\text{format}}$ denotes the format reward, $\lambda_1$ and $\lambda_2$ control the model's attention to regression precision and ranking accuracy, respectively, and $\lambda_3$ controls the trade-off between structured output compliance and task performance.

\subsection{Response Mutation Operation}
\label{sec:RMO}

A key challenge in policy optimization with GRPO is entropy collapse: the model can converge to a suboptimal deterministic policy. This occurs when a batch of generated responses is homogeneously incorrect, yielding zero advantage for all responses and causing the policy gradient to vanish. This traps the model in a saddle point, halting further learning.

To mitigate this limitation, we propose a Response Mutation Operation (RMO)---a targeted exploration mechanism inspired by evolutionary strategies. RMO probabilistically replaces $k$ low-reward responses in each training batch with ground-truth or high-reward reference answers. This intervention artificially introduces advantage variance ($\text{Var}(\hat{A}_i)>0$) by creating contrastive pairs between mutated and original responses, thereby reactivating gradient flow. Formally, for a mutated response $\boldsymbol{o}_j'$, its advantage is recalibrated as:
\begin{equation}
\begin{aligned}
\hat{A}_j'=\frac{R\left(\boldsymbol{q},\boldsymbol{o}_j'\right)-\text{mean}\left(\left\{r_i\right\}_{i=1}^G\right)}{\text{std}\left(\left\{r_i\right\}_{i=1}^G\right)}.
\end{aligned}
\end{equation}

This mechanism ensures robust policy updates, especially in ordinal regression tasks where categories are prone to confusion, thereby preventing learning from being hindered. 

We observed that simultaneous optimization of all three rewards from the outset leads to unstable training. To further stabilize training, we adopt a two-stage strategy. We first train the model on the regression task alone. In the second stage, we introduce the full ranking objective and activate RMO to jointly refine both regression and ranking capabilities.

\section{EXPERIMENTS}
\begin{table}[]
\caption{Performance comparison on UTKFace Dataset.}
\vspace{-10pt}
\label{tab:utkface}
\begin{tabular}{lcccc}
\shline
\multicolumn{1}{c}{\multirow{2}{*}{Methods}} & \multicolumn{4}{c}{UTKFace}                                                                                       \\
\multicolumn{1}{c}{}                         & \multicolumn{1}{l}{1-img$\downarrow$} & \multicolumn{1}{l}{2-imgs$\uparrow$} & \multicolumn{1}{l}{4-imgs$\uparrow$} & \multicolumn{1}{l}{8-imgs$\uparrow$} \\ \hline
MiVOLO                                      & 4.23                       & -                          & -                          & -                          \\
RankingCLIP                            & 3.83                       & -                          & -                          & -                          \\ \hline
Qwen2.5-VL-3B                                & 6.95                       & 0.712        &    0.704    & 0.693       \\ \hline
+ SFT                                        & 5.51                       & 0.771                         & 0.756                       & 0.748                          \\
+ RARL                                       & 4.02                       &  0.843                          &  0.831                         & 0.806                          \\ \hline
Qwen2.5-VL-7B                                & 6.14                       &            0.77                &            0.75                &      0.75                      \\ \hline
+ SFT                                        & 5.02                       & 0.783                          &  0.778                          & 0.774                          \\
+ RARL                                       & \textbf{3.81}              & \textbf{0.921}                         &  \textbf{0.905}                          & \textbf{0.852}                          \\ \shline
\end{tabular}
\vspace{-10pt}
\end{table}

\begin{table*}
\caption{Ablation studies on the different verifiable rewards. Notice that our method can eliminate the mandatory reliance on regression supervision and directly leverages sequence-level ranking results to guide the model's training.}
\vspace{-10pt}
\label{tab:rewards}
\begin{tabular}{lcccccccccccc}
\shline
\multicolumn{1}{c}{\multirow{2}{*}{Methods}} & \multicolumn{4}{c}{UTKFace}        & \multicolumn{4}{c}{COCO-REM}                                                                      & \multicolumn{4}{c}{AVA}            \\
\multicolumn{1}{c}{}                         & 1-img & 2-imgs & 4-imgs & 8-imgs & \multicolumn{1}{c}{1-img} & 2-imgs               & 4-imgs                & 8-imgs             & 1-img & 2-imgs & 4-imgs & 8-imgs \\ \hline
Baseline                                &   6.95    &   0.712     &   0.704     &     0.693     & 31.36                     &      0.744                 &       0.753                &       0.740                &    0.527   &     0.656       & 0.677       & 0.634    \\ \hline
+ Regression                         &   4.34    &    0.821    &    0.802     &     0.795      & 66.97                     & 0.802 & 0.782 & 0.779 &    0.766   &    0.821                          & 0.814                        & 0.808       \\
+ Ranking                            &     -  &     0.814      &    0.823      & 0.781                         &    -   &       0.804         &      0.797                 &            0.784           &   -    &   \textbf{0.858}     &     \textbf{0.842}   &    \textbf{0.827}     \\
+ RARL                                       &    \textbf{4.02}   &    \textbf{0.843}    &    \textbf{0.831}                         & \textbf{0.806}     & \textbf{68.73}                     & \textbf{0.831}                          & \textbf{0.824}                         & \textbf{0.797}  &   \textbf{0.783}    &    0.840                          & 0.832                          & 0.821      \\ \shline
\end{tabular}
\end{table*}

\begin{table}[]
\caption{Performance comparison (\%) on COCO-REM.}
\vspace{-10pt}
\label{tab:cocorem}
\begin{tabular}{lcccc}
\shline
\multicolumn{1}{c}{\multirow{2}{*}{Methods}} & \multicolumn{4}{c}{COCO-REM}                                                                                       \\
\multicolumn{1}{c}{}                         & \multicolumn{1}{l}{1-img$\uparrow$} & \multicolumn{1}{l}{2-imgs$\uparrow$} & \multicolumn{1}{l}{4-imgs$\uparrow$} & \multicolumn{1}{l}{8-imgs$\uparrow$} \\ \hline
\multicolumn{1}{c}{Qwen2.5-VL-3B}            & 31.36                     &  0.744      & 0.753       & 0.740         \\
+ SFT                                        & 66.97                     & 0.765                          & 0.774                          & 0.764                           \\
+ RARL                                       & 68.73                     & 0.831                          & 0.824                         & 0.797                            \\ \hline
\multicolumn{1}{c}{Qwen2.5-VL-7B}            & 39.46                     &            0.736                &               0.732             &      0.721                        \\
+ SFT                                        & 65.01                     & 0.745                          & 0.772                          & 0.763                           \\
+ RARL                                       & \textbf{71.80}            & \textbf{0.893}                          & \textbf{0.871}                        & \textbf{0.823}                            \\ \shline
\end{tabular}
\end{table}

\begin{table}[]
\caption{Performance comparison (\%) on AVA Dataset.}
\vspace{-10pt}
\label{tab:ava}
\begin{tabular}{lcccc}
\shline
\multicolumn{1}{c}{\multirow{2}{*}{Methods}} & \multicolumn{4}{c}{AVA}                                                                                           \\
\multicolumn{1}{c}{}                         & \multicolumn{1}{l}{1-img$\uparrow$} & \multicolumn{1}{l}{2-imgs$\uparrow$} & \multicolumn{1}{l}{4-imgs$\uparrow$} & \multicolumn{1}{l}{8-imgs$\uparrow$} \\ \hline
RankingCLIP                             & 0.747                      &            -                &      -                      &     -                       \\
VILA-R                                       & 0.774                      & -                          & -                          & -                          \\ \hline
Qwen2.5-VL-3B                                & 0.527                           & 0.656       & 0.677       & 0.634       \\ \hline
+ SFT                                        & 0.755                          & 0.770                          & 0.801                          & 0.806                          \\
+ RARL                                       & 0.783                       & 0.840                          & 0.832                          & 0.821                          \\ \hline
Qwen2.5-VL-7B                                & 0.643                           &         0.714                   &         0.723                   &      0.731                      \\ \hline
+ SFT                                        & 0.771                          & 0.815                         & 0.813                          & 0.823                          \\
+ RARL                                       & \textbf{0.803}                  & \textbf{0.875}                          & \textbf{0.883}                          & \textbf{0.871}                          \\ \shline
\end{tabular}
\vspace{-10pt}	
\end{table}

\subsection{Experimental Setup}

We evaluate RARL on three ordinal regression and ranking benchmarks: facial age estimation (UTKFace)~\cite{zhang2017age}, object counting (COCO-REM)~\cite{singh2024benchmarking}, and aesthetic assessment (AVA)~\cite{ghosal2019aesthetic}. For single-image regression (1-img), we report Mean Absolute Error (MAE) for age, accuracy (\%) for counting, and Spearman's Rank Correlation (SRCC) for aesthetics. For multi-image ranking (N-imgs), we use Kendall’s Tau ($\tau$). 

Our model is based on Qwen2.5-VL~\cite{bai2025qwen2}. We use a two-stage training strategy. Stage 1 focuses on regression ($\lambda_1=\lambda_3=1,\lambda_2=0$), while Stage 2 jointly optimizes for both regression and ranking ($\lambda_1=\lambda_2=\lambda_3=1$). Key hyperparameters include a learning rate of $1 \times 10^{-6}$ (AdamW), a batch size of $64$, and $k=2$ for RMO. The temperature for model generation is set to $1.0$, and KL divergence is set to $0.01$ to mitigate overfitting.

\subsection{Quantitative Results}

Table~\ref{tab:utkface}~\ref{tab:cocorem}~\ref{tab:ava} presents our main results. RARL consistently and significantly outperforms the base model and the Supervised Fine-Tuning (SFT) baseline across all tasks and settings. For instance, on the 7B model, RARL achieves a SOTA MAE of $3.81$ on UTKFace and SRCC of $0.803$ on AVA. The ranking performance ($\tau$) also shows substantial gains, especially in multi-image scenarios, demonstrating the effectiveness of our joint regression and ranking optimization. Although object counting is not a typical ordinal task, the performance boost is consistent even for this task, highlighting the benefits of RARL's multi-image comparative reasoning.


\begin{table}[]
\caption{Ablation studies on the effectiveness of the Two-stages Training and RMO on the UTKFace.}
\vspace{-10pt}
\label{tab:rmo}
\begin{tabular}{lcccc}
\shline
\multicolumn{1}{c}{\multirow{2}{*}{Methods}} & \multicolumn{4}{c}{UTKFace}                            \\
\multicolumn{1}{c}{}                         & \multicolumn{1}{c}{1-img$\downarrow$} & 2-imgs$\uparrow$ & 4-imgs$\uparrow$ & 8-imgs$\uparrow$ \\ \hline
Qwen2.5-VL-3B                                & 6.95                      & 0.712     &   0.704     &     0.693    \\ \hline
\multicolumn{5}{c}{First Stage}                                                                       \\ \hline
                                             & 4.34                      &     0.821     &   0.802     &     0.795       \\ \hline
\multicolumn{5}{c}{Second Stage}                                                                      \\ \hline
w.o. RMO                                     & 4.17                      &     0.832   &  0.825      &    0.803      \\
w. RMO                                       & \textbf{4.02}                      &     \textbf{0.843}   &    \textbf{0.831}                         & \textbf{0.806}     \\ \shline
\end{tabular}
\vspace{-10pt}	
\end{table}

\subsection{Ablation Studies}

\noindent\textbf{Impact of Reward Components.} As shown in Table~\ref{tab:rewards}, isolating the regression (+Reg) and ranking (+Rank) rewards both yield significant improvements over the baseline, but our full RARL model, which combines them, achieves the best performance. This confirms the synergistic effect of our unified objective. Notably, on subjective tasks like aesthetics, optimizing with ranking reward alone (+Rank) outperforms even the full regression model in ranking metrics (e.g., 0.858 vs. 0.821 on AVA 2-imgs). This suggests that for tasks with noisy labels, relative supervision can be more effective than absolute value supervision.

\noindent\textbf{Effect of two-stage training strategy.} As shown in Table~\ref{tab:rmo}, the two-stage training strategy effectively balances multiple rewards by training with different rewards in distinct stages. Adding RMO to the final stage consistently improves performance across both regression and ranking (e.g., MAE drops from $4.17$ to $4.02$). This validates RMO's role in overcoming policy stagnation and finding better optima.

\section{CONCLUSION}
This work introduces RARL, an efficient and scalable framework for ordinal ranking. By integrating regression and L2R objectives into a unified training paradigm and designing ranking-aware verifiable rewards, RARL synergistically enhances regression accuracy and ranking performance through bidirectional regularization—leveraging regression errors to refine ranking consistency and vice versa. To address entropy collapse in GRPO, we further propose RMO, which injects controlled perturbations into stagnant response groups to reactivate gradient signals and escape saddle points. Extensive experiments demonstrate state-of-the-art results across three benchmarks, with ablation studies revealing that ranking-centric training alone achieves robust ordinal performance, challenging conventional reliance on auxiliary regression supervision. Overall, our work underscores the viability of combining VLMs with RLVR, offering a principled pathway toward sample-efficient and interpretable ranking systems.
\bibliographystyle{IEEEbib}
\bibliography{strings,refs}

@String(AAAI = {AAAI})

@article{shen2017label,
  title={Label distribution learning forests},
  author={Shen, Wei and Zhao, Kai and Guo, Yilu and Yuille, Alan L},
  journal={Advances in neural information processing systems},
  volume={30},
  year={2017}
}

@article{rothe2018deep,
  title={Deep expectation of real and apparent age from a single image without facial landmarks},
  author={Rothe, Rasmus and Timofte, Radu and Van Gool, Luc},
  journal={International Journal of Computer Vision},
  volume={126},
  number={2},
  pages={144--157},
  year={2018},
  publisher={Springer}
}

@inproceedings{dai2023semi,
  title={Semi-supervised deep regression with uncertainty consistency and variational model ensembling via bayesian neural networks},
  author={Dai, Weihang and Li, Xiaomeng and Cheng, Kwang-Ting},
  booktitle={Proceedings of the AAAI Conference on Artificial Intelligence},
  volume={37},
  number={6},
  pages={7304--7313},
  year={2023}
}

@article{geng2013facial,
  title={Facial age estimation by learning from label distributions},
  author={Geng, Xin and Yin, Chao and Zhou, Zhi-Hua},
  journal={IEEE transactions on pattern analysis and machine intelligence},
  volume={35},
  number={10},
  pages={2401--2412},
  year={2013},
  publisher={IEEE}
}

@inproceedings{diaz2019soft,
  title={Soft labels for ordinal regression},
  author={Diaz, Raul and Marathe, Amit},
  booktitle={Proceedings of the IEEE/CVF conference on computer vision and pattern recognition},
  pages={4738--4747},
  year={2019}
}

@inproceedings{niu2016ordinal,
  title={Ordinal regression with multiple output cnn for age estimation},
  author={Niu, Zhenxing and Zhou, Mo and Wang, Le and Gao, Xinbo and Hua, Gang},
  booktitle={Proceedings of the IEEE conference on computer vision and pattern recognition},
  pages={4920--4928},
  year={2016}
}

@inproceedings{chen2017using,
  title={Using ranking-CNN for age estimation},
  author={Chen, Shixing and Zhang, Caojin and Dong, Ming and Le, Jialiang and Rao, Mike},
  booktitle={Proceedings of the IEEE conference on computer vision and pattern recognition},
  pages={5183--5192},
  year={2017}
}

@article{wang2023learning,
  title={Learning-to-rank meets language: Boosting language-driven ordering alignment for ordinal classification},
  author={Wang, Rui and Li, Peipei and Huang, Huaibo and Cao, Chunshui and He, Ran and He, Zhaofeng},
  journal={Advances in Neural Information Processing Systems},
  volume={36},
  year={2023}
}

@inproceedings{yu2024clip,
  title={CLIP-DR: Textual Knowledge-Guided Diabetic Retinopathy Grading with Ranking-Aware Prompting},
  author={Yu, Qinkai and Xie, Jianyang and Nguyen, Anh and Zhao, He and Zhang, Jiong and Fu, Huazhu and Zhao, Yitian and Zheng, Yalin and Meng, Yanda},
  booktitle={International Conference on Medical Image Computing and Computer-Assisted Intervention},
  pages={667--677},
  year={2024},
  organization={Springer}
}

@article{cheng2023robust,
  title={Robust image ordinal regression with controllable image generation},
  author={Cheng, Yi and Ying, Haochao and Hu, Renjun and Wang, Jinhong and Zheng, Wenhao and Zhang, Xiao and Chen, Danny and Wu, Jian},
  journal={arXiv preprint arXiv:2305.04213},
  year={2023}
}

@inproceedings{zhang2017age,
  title={Age progression/regression by conditional adversarial autoencoder},
  author={Zhang, Zhifei and Song, Yang and Qi, Hairong},
  booktitle={Proceedings of the IEEE conference on computer vision and pattern recognition},
  pages={5810--5818},
  year={2017}
}

@inproceedings{singh2024benchmarking,
  title={Benchmarking object detectors with coco: A new path forward},
  author={Singh, Shweta and Yadav, Aayan and Jain, Jitesh and Shi, Humphrey and Johnson, Justin and Desai, Karan},
  booktitle={European Conference on Computer Vision},
  pages={279--295},
  year={2024},
  organization={Springer}
}

@inproceedings{ghosal2019aesthetic,
  title={Aesthetic image captioning from weakly-labelled photographs},
  author={Ghosal, Koustav and Rana, Aakanksha and Smolic, Aljosa},
  booktitle={Proceedings of the IEEE/CVF International Conference on Computer Vision Workshops},
  pages={0--0},
  year={2019}
}

@article{yu2025dapo,
  title={Dapo: An open-source llm reinforcement learning system at scale},
  author={Yu, Qiying and Zhang, Zheng and Zhu, Ruofei and Yuan, Yufeng and Zuo, Xiaochen and Yue, Yu and Fan, Tiantian and Liu, Gaohong and Liu, Lingjun and Liu, Xin and others},
  journal={arXiv preprint arXiv:2503.14476},
  year={2025}
}

@article{liu2025understanding,
  title={Understanding r1-zero-like training: A critical perspective},
  author={Liu, Zichen and Chen, Changyu and Li, Wenjun and Qi, Penghui and Pang, Tianyu and Du, Chao and Lee, Wee Sun and Lin, Min},
  journal={arXiv preprint arXiv:2503.20783},
  year={2025}
}

@article{chu2025gpg,
  title={GPG: A Simple and Strong Reinforcement Learning Baseline for Model Reasoning},
  author={Chu, Xiangxiang and Huang, Hailang and Zhang, Xiao and Wei, Fei and Wang, Yong},
  journal={arXiv preprint arXiv:2504.02546},
  year={2025}
}

@inproceedings{kong2016photo,
  title={Photo aesthetics ranking network with attributes and content adaptation},
  author={Kong, Shu and Shen, Xiaohui and Lin, Zhe and Mech, Radomir and Fowlkes, Charless},
  booktitle={Computer Vision--ECCV 2016: 14th European Conference, Amsterdam, The Netherlands, October 11--14, 2016, Proceedings, Part I 14},
  pages={662--679},
  year={2016},
  organization={Springer}
}

@article{guo2025deepseek,
  title={Deepseek-r1: Incentivizing reasoning capability in llms via reinforcement learning},
  author={Guo, Daya and Yang, Dejian and Zhang, Haowei and Song, Junxiao and Zhang, Ruoyu and Xu, Runxin and Zhu, Qihao and Ma, Shirong and Wang, Peiyi and Bi, Xiao and others},
  journal={arXiv preprint arXiv:2501.12948},
  year={2025}
}

@article{team2025kimi,
  title={Kimi k1. 5: Scaling reinforcement learning with llms},
  author={Team, Kimi and Du, Angang and Gao, Bofei and Xing, Bowei and Jiang, Changjiu and Chen, Cheng and Li, Cheng and Xiao, Chenjun and Du, Chenzhuang and Liao, Chonghua and others},
  journal={arXiv preprint arXiv:2501.12599},
  year={2025}
}

@article{bai2025qwen2,
  title={Qwen2. 5-vl technical report},
  author={Bai, Shuai and Chen, Keqin and Liu, Xuejing and Wang, Jialin and Ge, Wenbin and Song, Sibo and Dang, Kai and Wang, Peng and Wang, Shijie and Tang, Jun and others},
  journal={arXiv preprint arXiv:2502.13923},
  year={2025}
}

\end{document}